\documentclass[conference]{IEEEtran}
\IEEEoverridecommandlockouts

\usepackage{cite}
\usepackage{amsmath,amssymb,amsfonts}
\usepackage{algorithmic}
\usepackage{graphicx}
\usepackage{textcomp}
\usepackage{xcolor}
\usepackage{comment}
\usepackage[hidelinks]{hyperref}
\usepackage{multirow}
\def\BibTeX{{\rm B\kern-.05em{\sc i\kern-.025em b}\kern-.08em
    T\kern-.1667em\lower.7ex\hbox{E}\kern-.125emX}}
\begin{document}

\title{A Multi-Camera Vision-Based Approach for Fine-Grained Assembly Quality Control\\

}


\author{
    \IEEEauthorblockN{Ali Nazeri\IEEEauthorrefmark{1}\IEEEauthorrefmark{2}, Shashank Mishra\IEEEauthorrefmark{1}\IEEEauthorrefmark{2}, Achim Wagner\IEEEauthorrefmark{2}, Martin Ruskowski\IEEEauthorrefmark{2}\IEEEauthorrefmark{3}, Didier Stricker\IEEEauthorrefmark{2}\IEEEauthorrefmark{3}, Jason Rambach\IEEEauthorrefmark{2}}
    \IEEEauthorblockA{\IEEEauthorrefmark{2}German Research Center for Artificial Intelligence (DFKI)\\
    \{ali.nazeri, shashank.mishra, achim.wagner, martin.ruskowski, didier.stricker, jason.rambach\}@dfki.de}
    \thanks{\IEEEauthorrefmark{1}Denotes equal contribution}
    \IEEEauthorblockA{\IEEEauthorrefmark{3}RPTU Kaiserslautern}
}


\maketitle

\begin{abstract}
Quality control is a critical aspect of manufacturing, particularly in ensuring the proper assembly of small components in production lines. Existing solutions often rely on single-view imaging or manual inspection, which are prone to errors due to occlusions, restricted perspectives, or lighting inconsistencies. These limitations require the installation of additional inspection stations, which could disrupt the assembly line and lead to increased downtime and costs. This paper introduces a novel multi-view quality control module designed to address these challenges, integrating a multi-camera imaging system with advanced object detection algorithms. By capturing images from three camera views, the system provides comprehensive visual coverage of components of an assembly process. A tailored image fusion methodology combines results from multiple views, effectively resolving ambiguities and enhancing detection reliability.
To support this system, we developed a unique dataset comprising annotated images across diverse scenarios, including varied lighting conditions, occlusions, and angles, to enhance applicability in real-world manufacturing environments. Experimental results show that our approach significantly outperforms single-view methods, achieving high precision and recall rates in the identification of improperly fastened small assembly parts such as screws. This work contributes to industrial automation by overcoming single-view limitations, and providing a scalable, cost-effective, and accurate quality control mechanism that ensures the reliability and safety of the assembly line. The dataset used in this study is publicly available to facilitate further research in this domain. It can be accessed at \url{https://cloud.dfki.de/owncloud/index.php/s/CkCHqbwPjMCsiQf}.

\end{abstract}


\section{Introduction}
 In modern manufacturing, maintaining product quality is essential, as defects can result in customer dissatisfaction, elevated costs, and safety risks. Conventional manual inspection techniques tend to be labor-intensive and susceptible to errors, which makes them inadequate for the high standards in modern production settings. Studies indicate high error rates associated with manual inspections, highlighting the need for more reliable solutions \cite{r01}. To address these challenges, machine vision systems have emerged as an innovative solution, enhancing the automation of quality control processes with enhanced accuracy and efficiency. These systems employ advanced machine vision methods and algorithms to identify defects that may be missed by human inspectors, thus improving product quality and consistency \cite{r02}.
 
 Machine vision contains the combination of cameras, sensors, and tailored software to capture and analyze visual data, which can be employed for automated inspection and quality control. Automated machine vision systems provide numerous advantages compared to manual solutions, including enhanced speed, consistency, and the capability to identify even small errors that may go unrecognized by human observers. In the aerospace applications, machine vision plays an important role in specifying surface defects and inspecting assemblies, leading to enhanced safety and reliability of components \cite{r03}. Although traditional machine vision systems employing a single camera offer certain benefits, they often encounter limitations when faced with complex situations like components featuring complicated geometry or small assembly components. Challenges like occlusions, various lighting conditions, and restricted viewing angles can impede precise inspection. To address these challenges, advanced multi-camera inspection technologies have been proposed to ensure comprehensive coverage and facilitate the development of digital twins for further analysis \cite{r04}. The combination of multi-camera systems and advanced image processing algorithms has resulted in notable enhancements in quality control procedures. By capturing multiple perspectives, these systems can effectively address ambiguities that arise in single-view inspections, resulting in more precise defect detection and an overall decrease in error rates \cite{r04}.\\
Machine vision in quality control faces challenges like lighting consistency, camera calibration, and task-specific algorithms. Advancements in machine learning have enabled more robust systems that improve accuracy with data-driven learning, reducing the need for manual programming \cite{r05}.

This study focuses on the challenge of implementing vision-based quality control for the assembly of small components on industrial production lines, with a specific emphasis on screw tightness control. In order to accomplish this, we gathered and labeled a dataset that had not been accessible before for this task, and we introduce a method for object detection that integrates multi-view information.

\section{Related Work}
Multi-camera vision systems are emerging into an effective approach to tackle the complexity of dynamic environments. By means of the collection of images from various angles, they address challenges such as occlusions, limited viewpoints, and variations in lighting conditions\cite{r06}.

A recent investigation introduced a multi-camera vision system proposed for the automated measurement of automotive components. This system offers the capability to demonstrate accurate dimensional control and providing real-time inspection in high-accuracy industrial applications \cite{r06}. A comprehensive framework for multi-view defect detection has been developed to tackle the challenges of complex object detection in the printed circuit boards (PCBs) inspection process. The system demonstrated a significant progress in classification accuracy, especially for small components, through the incorporation of multiple viewpoints into an effective decision-making framework \cite{r07}.

Dehaerne et al. stated that the implementation of semi-automatic ground-truth generation has enhanced detection reliability, resulting in a notable improvement in mean average precision (mAP) when compared to single-view methods for the object detection task \cite{r08}.

Object detection employing deep learning techniques has significantly contributed to the progress of automated assembly inspection. YOLOv8 \cite{b12} and ByteTrack \cite{b14} were incorporated into real-time tracking systems for monitoring the errors in the assembly process in manufacturing. This method enabled precise identification and monitoring of failures, while simultaneously reducing the need for human involvement \cite{r09}. The implementation of multi-camera fusion techniques greatly improved the inspection of rolling mills through the analysis of visual sequences captured from various angles. Automation driven by machine vision effectively detected errors in rolling stock, which leads to minimized downtime and enhanced quality control practices \cite{r10}.

In the field of process control, a cross-machine control loop was established to improve production ramp-up by dynamically modifying process parameters in response to real-time machine data. Random forest models were employed to forecast upstream and downstream effects, thereby ensuring consistency in the product quality \cite{r11}.\\
In medical equipment production, where the accuracy and quality control are critical, ML models for anomaly detection utilizing one-class support vector machines (SVMs) and binary classifiers have successfully identified defective products with remarkable accuracy. This provides an effective approach to addressing the challenges posed by limited defect samples \cite{r12}. These machine learning models facilitate adaptive quality control and minimize error rates.
Taking advantage of structured light and projected texture stereo vision has greatly enhanced the reliability of defect detection and the localization of components. Through the projection of controlled light patterns onto various objects, these techniques produce high-resolution 3D surface reconstructions which enhance inspection accuracy even in environments lacking texture \cite{r13}. In the field of aerospace manufacturing, a mobile collaborative robot, utilizing structured blue light sensors, autonomously identified and located parts, thereby decreasing the dependence on manual inspection \cite{r14}. The combination of structured light and multi-camera vision systems is a promising approach for significantly improving defect detection by reducing occlusions and enhancing depth estimation in the quality control process \cite{r15}.
\section{Methodology}

\subsection{Dataset Collection and Annotation}
A major challenge in the development of vision-based quality control systems is the lack of a high-quality dataset specifically designed for industry-grade screw/bolt detection. Existing datasets are either too generic or lack the necessary diversity to cover real-world manufacturing conditions. To address this gap, we created a dedicated dataset using a three-camera setup (Fig~\ref{fig:camera_setup}), ensuring comprehensive coverage of screw/bolt assemblies on industrial frames. Our setup consisted of:

\begin{figure}[htbp]
    \centering
    \includegraphics[width=0.45\textwidth]{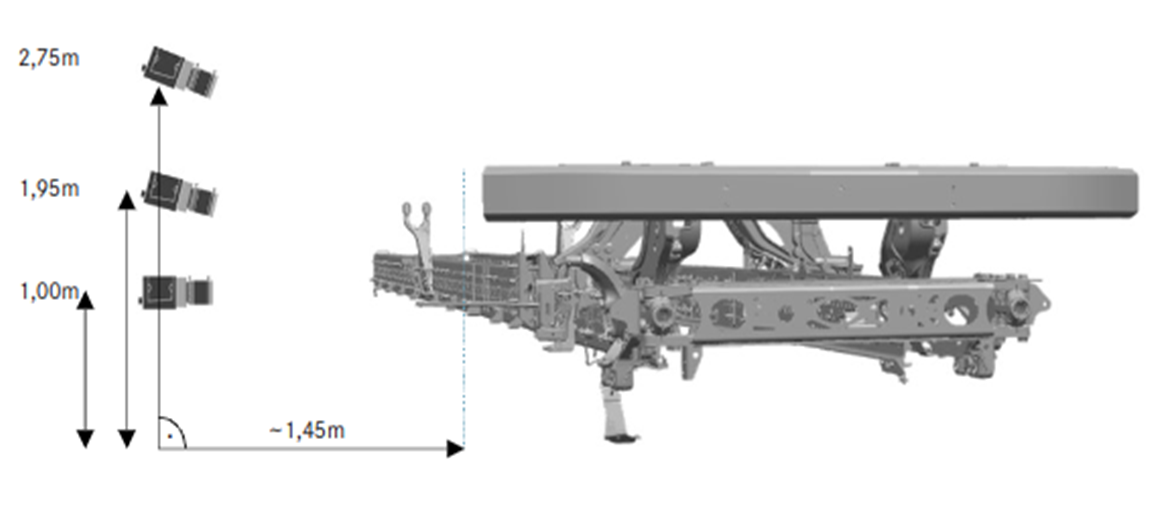}
    \caption{Camera Setup for Data Capturing.}
    \label{fig:camera_setup}
\end{figure}

\begin{enumerate}
    \item \textbf{Top Camera:} Capturing the near-side frame from a top-down view.
    \item \textbf{Middle Camera:} Focusing on the far-side frame, covering details that might be obscured from other angles.
    \item \textbf{Bottom Camera:} Capturing the same near-side frame but from a front-facing perspective.
    
\end{enumerate}

\begin{figure}[htbp]
    \centering
    \includegraphics[width=0.45\textwidth]{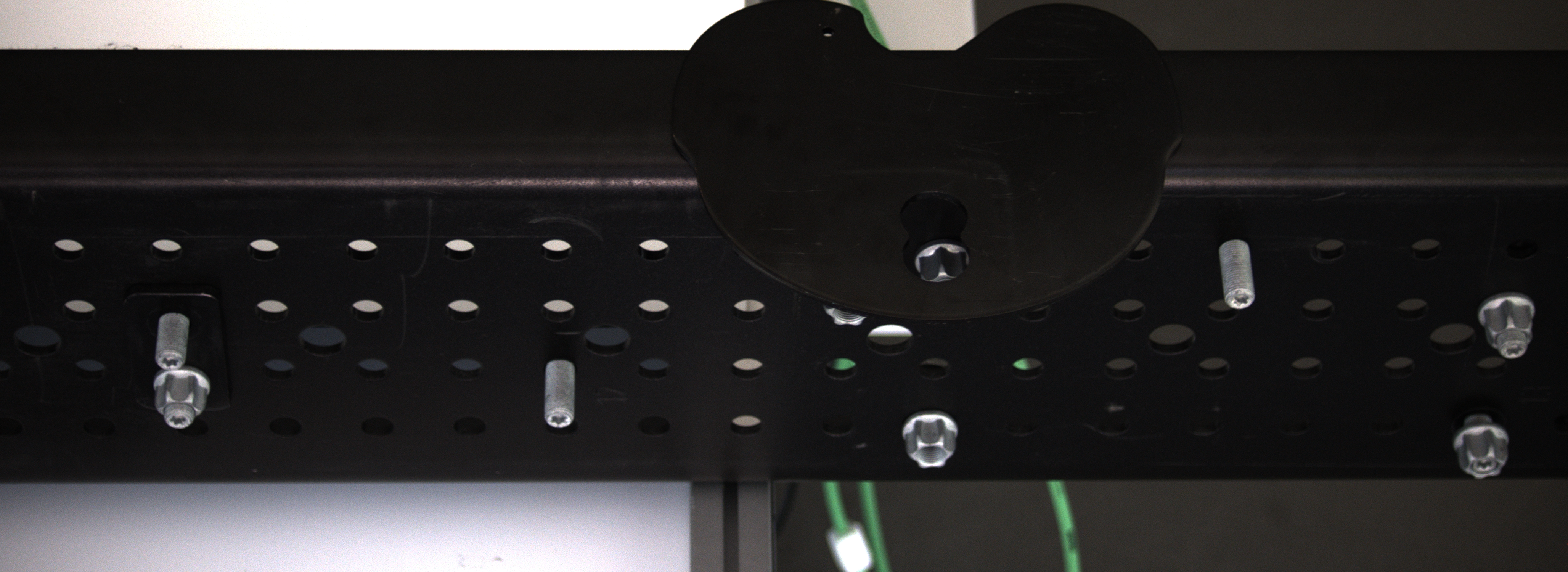}
    \includegraphics[width=0.45\textwidth]{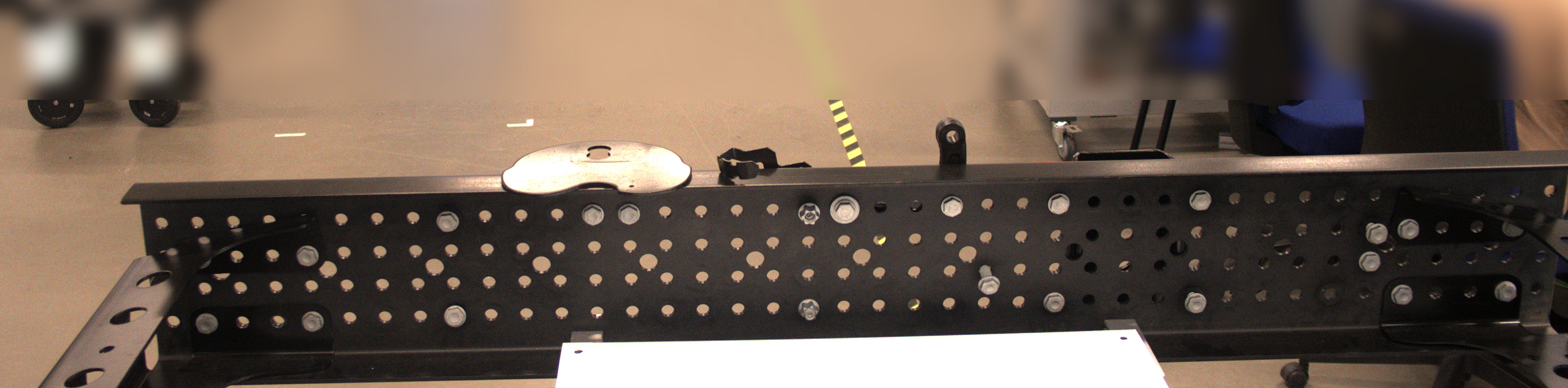}
    \includegraphics[width=0.45\textwidth]{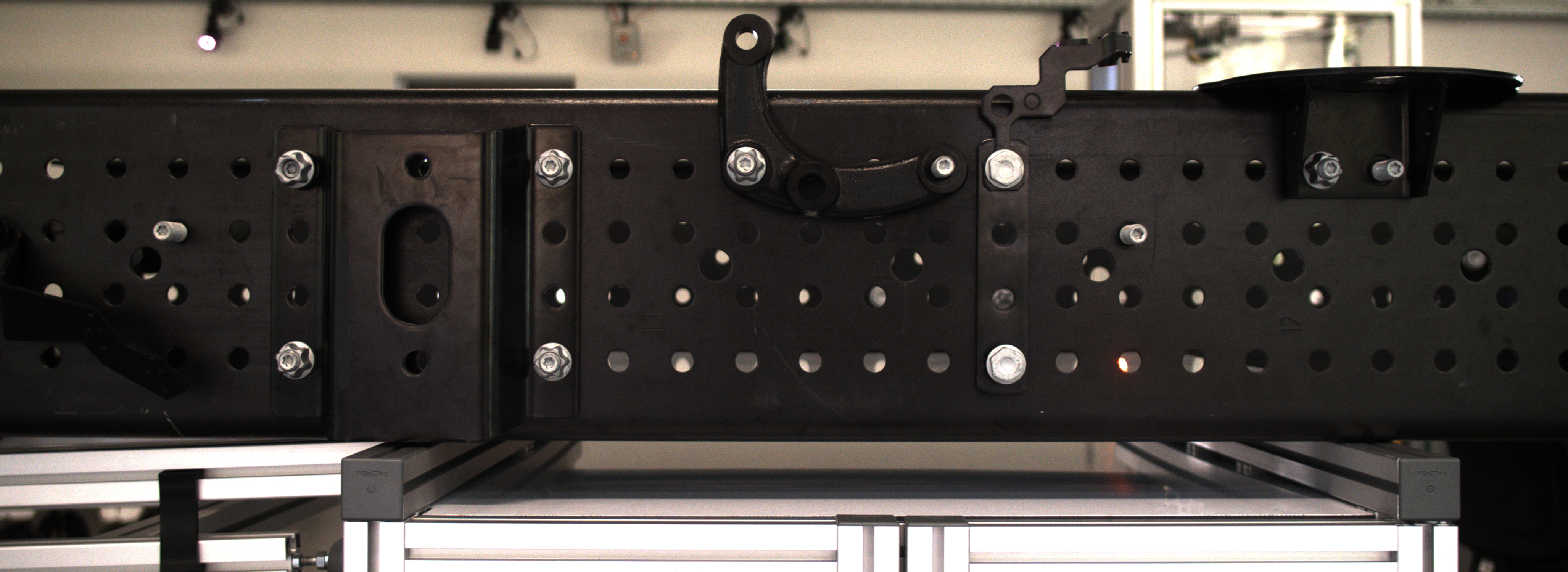}
    \caption{Top, Middle and Bottom camera images, encompassing both positive and negative cases.}
    \label{fig:camera_samples}
\end{figure}

Using this setup, we collected 1,200 images (400 per camera) under various conditions (Fig~\ref{fig:camera_samples}). From this set, we carefully curated 600 images that best represented real-world scenarios, considering factors such as occlusions, lighting variations, and different screw/bolt tightness levels, including gap variations of a few millimeters between the bolt and the frame. Each of these images was meticulously annotated to provide a high-quality, industry-standard benchmark dataset for screw/bolt detection. For model development, we employed a 70-15-15 stratified split, allocating 70\% of the dataset for training, 15\% for validation, and 15\% for testing. The partitioning process ensured an equal distribution of images across all three camera perspectives, maintaining a balanced representation of different viewpoints. 

This dataset is one of the first of its kind, specifically designed for industrial applications, ensuring robust model evaluation and deployment in real-world manufacturing environments.

\subsection{Model Selection and Optimization}
To develop an efficient and deployable screw/bolt detection system, we first explored various machine learning and deep learning approaches used in industrial inspection tasks. While numerous object detection methods exist \cite{r16, r17, r18, r19}, we found that most were either not specifically designed for this use case or lacked real-time inference capabilities required for industrial deployment.

The primary constraint in our setting is that the industrial pipeline (e.g., conveyor system) operates at a constant speed, and the quality inspection must be performed in real-time without altering the production flow or introducing external interference. This required a model capable of high-speed inference while maintaining state-of-the-art detection accuracy.

\subsubsection{Evaluating Different Architectures}
To explore potential architectures for the task, we reviewed several deep learning models, including:
\begin{itemize}
    \item Traditional CNN-based object detection models (e.g., Faster R-CNN \cite{r21}, SSD \cite{r22}) have been shown to offer good accuracy but may struggle with inference speed, which could be a limitation for real-time applications.
    \item Lightweight detection models (e.g., MobileNet-SSD \cite{r23}, EfficientDet \cite{r24}) are known for faster inference times, but they often face challenges in generalizing to difficult cases such as occlusions, varying lighting conditions, and different bolt tightness levels.
    \item Transformer-based detectors (e.g., DETR \cite{r25}, Deformable DETR \cite{r26}) provide strong feature extraction capabilities, but their high computational demands may hinder their suitability for real-time deployment in industrial settings.
\end{itemize}

Based on our review of existing models, YOLOv8 \cite{b12} emerged as a strong candidate for industrial applications, offering a balance between accuracy, speed, and efficiency. Its anchor-free detection mechanism and enhanced feature aggregation have been shown to improve performance, while its lightweight architecture enables real-time inference on industrial-grade hardware.

\subsubsection{Fine-Tuning YOLOv8 \cite{b12} for Industrial Deployment}
To further optimize the model for our specific task, we fine-tuned key hyperparameters based on iterative experimentation. We set the learning rate to 0.001, utilizing a cyclical learning rate strategy to achieve a balance between convergence speed and stability. The batch size was chosen as 8, considering memory constraints while ensuring stable gradient updates. We trained the model for 50 epochs, which was sufficient to reach convergence without overfitting. For optimization, we employed AdamW \cite{b13}, as it provides stable weight updates and enhances generalization. The loss function was designed to incorporate Cross-Entropy Loss for screw/bolt classification and IoU Loss for bounding box regression, ensuring precise localization of screw/bolt positions.

This comprehensive evaluation and fine-tuning process ensured that YOLOv8 \cite{b12} could deliver state-of-the-art accuracy while maintaining the real-time performance necessary for seamless industrial deployment.

\section{Results and Discussion}
To evaluate the effectiveness of our approach for detecting screw and bolt tightness/looseness, we conducted a series of experiments using a multi-camera setup shown in Fig~\ref{fig:camera_setup}. We trained and tested the YOLOv8-Large (YOLOv8l) \cite{b12} object detection model on images from all three cameras, using our selected hyperparameters. The evaluation metrics include Precision, Recall, mAP@50, mAP@50-95, and Inference Speed (measured in milliseconds).

\subsection{Multi-Camera Training and Performance}
We trained YOLOv8l \cite{b12} using both pretrained weights (initialized with COCO-trained \cite{r20} weights) and random initialization (trained from scratch). Table~\ref{tab:yolov8_results} presents the quantitative results.

\begin{table}[htbp]
    \centering
    \caption{Performance Comparison of YOLOv8l \cite{b12}}
    \resizebox{\columnwidth}{!}{
    \begin{tabular}{|c|c|c|c|c|c|c|}
        \hline
        \multirow{2}{*}{\textbf{Model}} & \multirow{2}{*}{\textbf{Pretrained}} & \multirow{2}{*}{\textbf{Precision}} & \multirow{2}{*}{\textbf{Recall}} & \textbf{mAP} & \textbf{mAP} & \textbf{Inference} \\
        & & & & \textbf{@50} & \textbf{@50-95} & \textbf{(ms)} \\
        \hline
        YOLOv8l \cite{b12} & True  & \textbf{1.000} & \textbf{0.999} & \textbf{0.995} & \textbf{0.994} & \textbf{9.0} \\
        \hline
        YOLOv8l \cite{b12} & False & 0.970 & 0.962 & 0.967 & 0.924 & 8.8 \\
        \hline
    \end{tabular}
    }
    \label{tab:yolov8_results}
\end{table}

Our results demonstrate that the pretrained YOLOv8l \cite{b12} model achieves state-of-the-art performance, with near-perfect detection accuracy. The model trained from scratch (without pretrained weights) shows a slight drop in performance, particularly in mAP@50-95, indicating the importance of transfer learning for this task. The mAP@50 and mAP@50-95 scores confirm that the model generalizes well across different screw/bolt placements and camera perspectives.

\subsection{Individual Camera Training and Evaluation}
In this experiment, we trained the YOLOv8l \cite{b12} model separately on images from each camera—Top, Middle, and Bottom—and subsequently evaluated the performance on the corresponding test sets. This camera-specific evaluation is crucial as it allows us to pinpoint any performance discrepancies that might arise due to unique characteristics inherent to each camera’s perspective. By isolating the data from each camera, we can assess the robustness of our model in different real-world conditions and determine whether tailored adjustments are required for optimal performance across all viewpoints.
\begin{table}[htbp]
    \centering
    \caption{Performance Metrics for Individual Camera Experiments}
    \resizebox{\columnwidth}{!}{
    \begin{tabular}{|c|c|c|c|c|c|c|c|c|}
        \hline
        \multirow{3}{*}{\textbf{Camera}} & \multicolumn{4}{c|}{\textbf{Validation Metrics}} & \multicolumn{4}{c|}{\textbf{Test Metrics}} \\
        \cline{2-9}
         & \multirow{2}{*}{\textbf{Precision}} & \multirow{2}{*}{\textbf{Recall}} & \textbf{mAP} & \textbf{mAP} & \multirow{2}{*}{\textbf{Precision}} & \multirow{2}{*}{\textbf{Recall}} & \textbf{mAP} & \textbf{mAP} \\
         & & & \textbf{@50} & \textbf{@50-95} & & & \textbf{@50} & \textbf{@50-95} \\
        \hline
        Top    & 0.956 & 0.965 & 0.989 & 0.806 & 0.900 & 0.987 & 0.959 & 0.754 \\
        \hline
        Middle & 0.945 & 0.242 & 0.303 & 0.227 & 0.971 & 0.241 & 0.289 & 0.217 \\
        \hline
        Bottom & 0.914 & 0.665 & 0.789 & 0.704 & 0.866 & 0.658 & 0.721 & 0.643 \\
        \hline
    \end{tabular}
    }
    \label{tab:individual_camera_results}
\end{table}

From Table \ref{tab:individual_camera_results}, we observe that the model performs robustly on the Top camera, achieving high precision and recall as well as excellent mAP scores on both the validation and test sets. However, for the Middle and Bottom cameras, the performance is significantly lower—especially for the Middle camera, where both recall and mAP values are substantially reduced. These discrepancies underscore the importance of individual camera testing, as they highlight potential issues such as variations in illumination, resolution, or viewpoint that might require further investigation or tailored enhancements to the model pipeline.

\subsection{Cross-Camera Testing}
To further evaluate the generalizability of the model across different viewpoints, we conducted cross-camera testing. In this experiment, the YOLOv8l \cite{b12} model was trained on images from one camera and tested on another. This analysis helps assess whether features learned from one camera's perspective can be effectively transferred to another viewpoint. The results of this experiment are summarized in Table \ref{tab:cross_camera_results}:

\begin{table}[htbp]
    \centering
    \caption{Cross-Camera Testing Performance}
    \resizebox{\columnwidth}{!}{
    \begin{tabular}{|c|c|c|c|c|c|}
        \hline
        {\textbf{Train}} & {\textbf{Test}} & \multirow{2}{*}{\textbf{Precision}} & \multirow{2}{*}{\textbf{Recall}} & \textbf{mAP} & \textbf{mAP} \\
        \textbf{Camera} & \textbf{Camera} & & & \textbf{@50} & \textbf{@50-95} \\
        \hline
        Top    & Middle & 0.006  & 0.035  & 0.005  & 0.001  \\
        \hline
        Top    & Bottom & 0.261  & 0.320  & 0.2411 & 0.157  \\
        \hline
        Middle & Top    & 0.000  & 0.010  & 0.000  & 0.000  \\
        \hline
        Middle & Bottom & 0.012  & 0.210  & 0.014  & 0.005  \\
        \hline
        Bottom & Top    & 0.374  & 0.321  & 0.124  & 0.067  \\
        \hline
        Bottom & Middle & 0.4412 & 0.294  & 0.119  & 0.079  \\
        \hline
    \end{tabular}
    }
    \label{tab:cross_camera_results}
\end{table}

The Table \ref{tab:cross_camera_results} shows, that the model struggles to generalize when trained on one camera and tested on another. The precision, recall, and mAP scores are significantly lower compared to previous experiments in which the model was trained and tested on the same camera. 

These results highlight the necessity of using all three camera views together for model training. The significant drop in performance suggests that each camera captures unique perspectives, making feature transfer challenging. Factors like angle, lighting, occlusions, and focus variations contribute to these discrepancies.

\subsection{Two-Camera Training with Cross-Camera Testing}
To further emphasize the necessity of a three-camera setup, we conducted an additional experiment where the model was trained using images from two cameras and tested on the third, unseen camera. This setup evaluates whether combining two viewpoints is sufficient for generalization to the remaining perspective. The results are summarized in Table \ref{tab:two_camera_cross_test}.

\begin{table}[htbp]
    \centering
    \caption{Performance of Two-Camera Training with Cross-Camera Testing}
    \resizebox{\columnwidth}{!}{
    \begin{tabular}{|c|c|c|c|c|c|}
        \hline
        {\textbf{Train}} & {\textbf{Test}} & \multirow{2}{*}{\textbf{Precision}} & \multirow{2}{*}{\textbf{Recall}} & \textbf{mAP} & \textbf{mAP} \\
        \textbf{Cameras} & \textbf{Camera} & & & \textbf{@50} & \textbf{@50-95} \\
        \hline
        Top-Bottom    & Middle & 0.219  & 0.624  & 0.212  & 0.155  \\
        \hline
        Top-Middle    & Bottom & 0.205  & 0.508  & 0.210  & 0.152  \\
        \hline
        Middle-Bottom & Top    & 0.475  & 0.344  & 0.210  & 0.126  \\
        \hline
    \end{tabular}
    }
    \label{tab:two_camera_cross_test}
\end{table}

From the Table \ref{tab:two_camera_cross_test}, we observe that while training on two cameras slightly improves performance compared to single-camera training, the model still struggles to generalize to an unseen camera. Precision, recall, and mAP scores remain significantly lower than those obtained when training on all three cameras. Notably, the Middle camera again presents challenges, with the model achieving only 0.219 precision and 0.212 mAP@50 when trained on the Top and Bottom cameras. This suggests that critical spatial or appearance-based features unique to the Middle camera are missing when it is excluded from training.

These findings reinforce the necessity of incorporating all three cameras during training. Even with two cameras providing partial coverage, the model lacks the full range of viewpoints needed for robust screw/bolt detection. This experiment justifies the importance of a multi-camera setup, as training on all three perspectives enables the model to learn a more comprehensive feature representation, significantly improving generalization across different viewpoints in an industrial setting.

\subsection{Real-World Evaluation}
To assess the model’s performance in a real production environment, we tested our fully trained YOLOv8l \cite{b12} model on images captured during actual manufacturing operations. These images were not part of the training or validation sets, providing an unbiased evaluation of the model’s robustness. The model performed as expected, accurately detecting screw/bolt conditions across various real-world scenarios. While no quantitative metrics were recorded for this test, qualitative results confirm that the model generalizes well to unseen images. Sample detections are shown below.

\begin{figure}[htbp]
    \centering
    \includegraphics[width=0.48\textwidth]{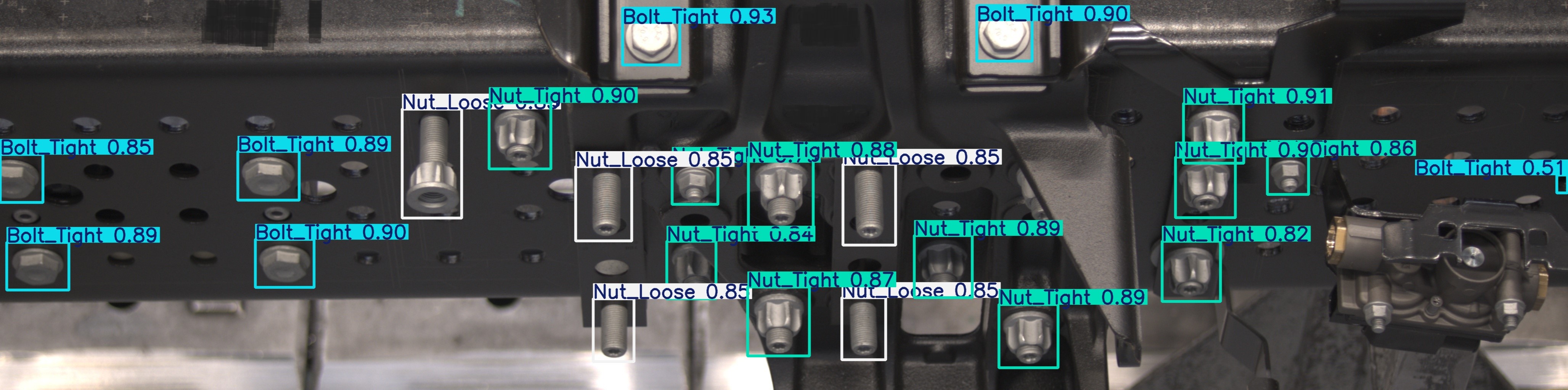}
    \includegraphics[width=0.48\textwidth]{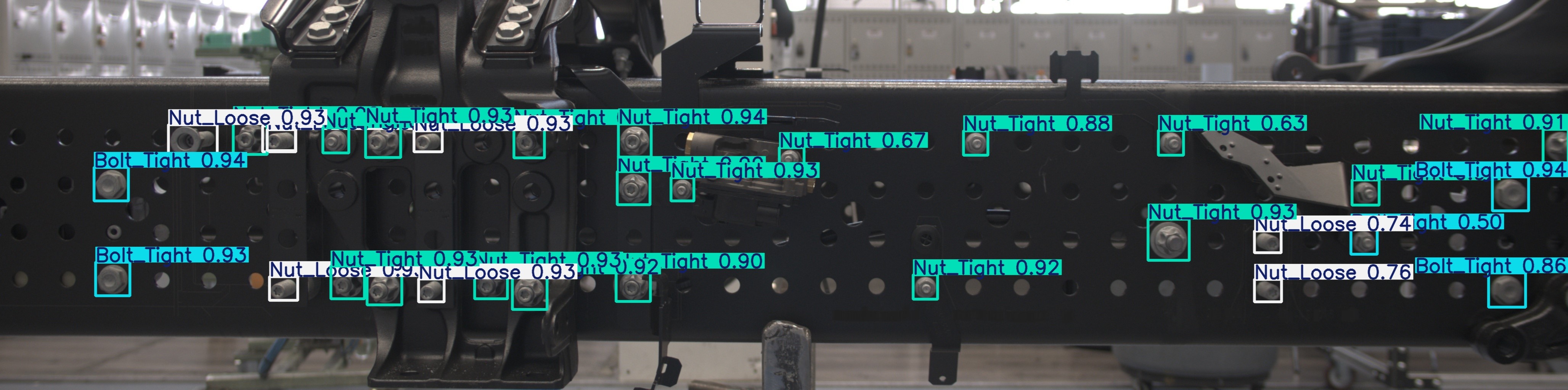}
    \includegraphics[width=0.48\textwidth]{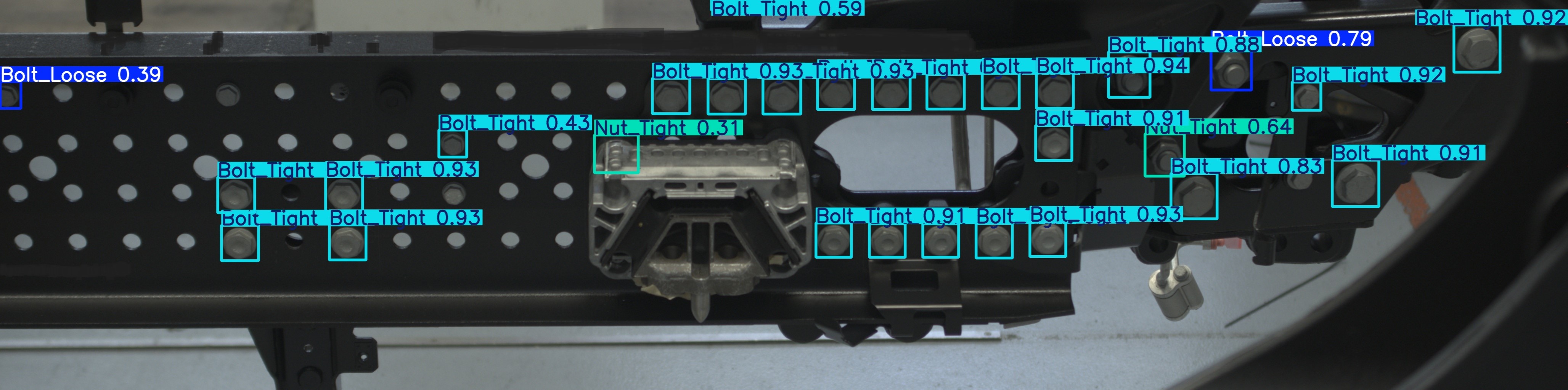}
    \caption{Qualitative results of the trained model on real-world production images, demonstrating its robustness in detecting screw/bolt conditions.}
    \label{fig:qualitative_results}
    \vspace{-3mm}
\end{figure}

\section{Conclusion}
We presented a multi-camera vision-based approach for fine-grained assembly quality control, focusing on the critical task of screw and bolt tightness inspection. As screws and bolts are the most widely used fastening components across industries, ensuring their proper installation is essential for product reliability and safety. Our multi-camera setup addresses occlusion and lighting challenges, providing a more reliable inspection compared to single-view methods. Our evaluation showed that YOLOv8 \cite{b12} delivers near-perfect detection performance, achieving a mAP@50 of 0.995 with an inference speed of 9.0 ms, making it highly suitable for real-time deployment. The proposed multi-camera training strategy further enhances detection accuracy by leveraging multiple viewpoints, minimizing inspection blind spots.

Future work will focus on expanding the approach to other assembly components and integrating temporal analysis for improved defect detection. This research contributes to more efficient and reliable manufacturing quality control processes.

\section*{Acknowledgements}
This work was partially funded by the German
Ministry for Economics and Climate Action (BMWK) under Grant Agreement 13IK010 (TWIN4TRUCKS).

\end{document}